\documentclass[mlmain,twocolumn]{jmlr}

\usepackage{rotating}
\usepackage{longtable}

\usepackage{booktabs}
\usepackage[load-configurations=version-1]{siunitx} 

\theorembodyfont{\upshape}
\theoremheaderfont{\scshape}
\theorempostheader{:}
\theoremsep{\newline}

\title[Activity Recognition in Surgical Videos]{An Empirical Study on Activity Recognition\titlebreak in Long Surgical Videos}

  \author{\Name{Zhuohong He} \Email{zooey.he@intusurg.com}\\
  \Name{Ali Mottaghi} \Email{mottaghi@stanford.edu}\\
  \Name{Aidean Sharghi} \Email{aidean.sharghikarganroodi@intusurg.com}\\
  \Name{Muhammad Abdullah Jamal} \Email{Abdullah.Jamal@intusurg.com}\\
  \Name{Omid Mohareri} \Email{omid.mohareri@intusurg.com}\\
\addr 1020 Kifer Rd. Sunnyvale, CA 94086}

\begin{document}

\maketitle

\begin{abstract}
Activity recognition in surgical videos is a key research area for developing next-generation devices and workflow monitoring systems. Since surgeries are long processes with highly-variable lengths, deep learning models used for surgical videos often consist of a two-stage setup using a backbone and temporal sequence model. In this paper, we investigate many state-of-the-art backbones and temporal models to find architectures that yield the strongest performance for surgical activity recognition. We first benchmark the models performance on a large-scale activity recognition dataset containing over 800 surgery videos captured in multiple clinical operating rooms. We further evaluate the models on the two smaller public datasets, the Cholec80 and Cataract-101 datasets, containing only 80 and 101 videos respectively. We empirically found that Swin-Transformer+BiGRU temporal model yielded strong performance on both datasets. Finally, we investigate the adaptability of the model to new domains by fine-tuning models to a new hospital and experimenting with a recent unsupervised domain adaptation approach.

\end{abstract}

\section{Introduction}
\label{sec:intro}

Exacerbated by nation-wide staffing shortages, hospitals are focusing more on improving efficiency across all aspects of operation. No where is this efficiency more linked to patient outcomes than the operating room (OR), where time under anesthesia, staff fatigue, and case turnover directly impact the quality and quantity of care. Additionally, OR efficiency can improve hospital profits and increase staff satisfaction and retention. Thus, a growing research field is working to model and analyze OR efficiency. Within the field, activity recognition of long surgical videos is a core technology which uses computer vision and deep learning to identify the activities or phases to generate a surgery timeline. Due to the ubiquity of camera sensors, surgical videos can be quickly collected and analyzed at scale.

Two of the most information-rich types of surgical videos are the videos captured by surgical tools (endoscopes, microscopes) and the external videos capturing the OR environment. Understanding endoscopic videos of laparoscopic and robotic surgery cases give us insight into surgeon efficiency and skill. External videos of the busy OR theatre allows us to understand how the idiosyncrasies of the hospital staff and OR might affect the surgery workflow and outcome. We denote this task as Operating Room Activity Recognition. Recently, advancements within the computer vision field have improved the state-of-the-art on long video segmentation; however, their suitability is untested on surgical videos.

Activity recognition in surgical videos presents several unique challenges over general activity recognition. First, surgical videos are long, resulting in the need to model long-reaching dependencies between activities often separated by 3 or more hours. The activities themselves range from several minutes to several hours in length. Second, surgeries are complex processes, requiring annotations from niche medical experts. Consequently, the datasets are often small, suffer from class imbalances, and contain complex patterns. Thus, it is hard to train video understanding models on these diverse datasets and the trained models don't usually generalize.

To model long-term dependencies, most existing works use a two-stage training scheme. The backbone model is first trained to extract features from sub-units of the video. A second temporal sequence model utilizes the features to generate activity predictions with long-term dependencies. Recent approaches for solving surgical activity recognition~\citep{endonet,tecno} use a frame-wise backbone model which focuses solely on creating ``spatial" representations through single frames, leaving the ``temporal" learning to the second stage. However, we find that the state-of-the-art activity recognition models~\cite{tformer,slowfast,swin} utilize short video clips (16 or 32 frames) to extract global representations. 

The goal of this work is to demonstrate the capabilities of spatial-temporal models on three surgical video datasets. First, we benchmarked the models on a new large-scale dataset we call the \textit{OR-AR} (Operating Room Activity Recognition) which was first introduced by \citet{mic20} and was later extended from 400 videos to 820 videos. Second, we show our model's state-of-the-art performance on two smaller public surgical datasets: Cholec80 \citep{endonet} and Cataracts-101 \citep{cataract101-dataset}) refuting the perception that spatial-temporal models perform poorly on small datasets. We delineate the best spatial-temporal backbone/temporal model combination and analyze the models' generalizability through experiments on challenging data splits and adaptation to new domains. 

Our main findings and contributions are as follows:
\begin{itemize}
  \item To the best of our knowledge, this is the first extensive evaluation of surgical activity recognition using different deep learning architectures.
  \item We find that Swin-Transformer + Bidirectional GRU is the best performing two-stage model. This model forms the new state-of-the-art on Cholec80 by achieving +2.32\% accuracy, +0.35\% precision, and +3.44\% recall when compared to the previous best. The model performs similarly well on the OR-AR dataset, and nearly matches the SOTA performance on Cataract-101 without additional processes such as contrastive learning or increased supervision.
  \item We find that I3D backbone is the most efficient model achieving 98.9\% of the Swin-Transformer's performance on OR-AR using ~66\% less parameters and ~80\% less FLOPs.
  \item We used an unsupervised technique to adapt our model to a new domain, a new hospital. Using an unseen hospital's dataset, which is 8\% the size of OR-AR, we achieved 93\% of the performance (on OR-AR) with no labels. Using labels, we achieved 96\% of the performance.

\end{itemize}

\section{Related Work}
In this section, we mostly review the current trends and approaches for surgical activity recognition.

Surgical activity recognition on \textit{laparoscopic videos} has been extensively studied in the literature thanks to well-made public datasets such as the Cholec80, first introduced by \citet{endonet}. \citet{endonet} also developed a multi-task learning approach to extract features from cholecystectomy video frames that capture both low-level visual information as well as tool-presence signals. \citet{cnnlstm} replaced the traditional HMMs with recurrent neural networks (RNN). \citet{mtrcn} proposed a correlation loss to capture the relevance between tool detection and activity recognition tasks to enrich the representation learning and utilized an RNN to learn the temporal dependencies between frames. TeCNO \citep{tecno} used temporal convolution networks \citep{tcn} on top of frame-wise features generated using ResNet to achieve state-of-the-art performance on Cholec80 and Cholec51 datasets. 

On the other hand, surgical activity recognition in \textit{robotic OR videos} was first introduced by \citet{mic20}, which collected the large-scale OR activity recognition dataset (OR-AR) and developed a supervised model which can recognize the start and end of significant steps. They followed a two-stage training where I3D \citep{i3d}, a 3D convolution-based model, was first pretrained and used for clip-wise feature extraction. Then, they trained a LSTM \citep{lstm} using extracted features to model the global dependencies between different activities.~\citet{mic21} introduced a multi-view model using the same building blocks. \citet{mic22} proposed a multi-modal unsupervised approach for low-data regime.

Finally, surgical activity recognition in \textit{cataract surgery videos} is another burgeoning research area benefiting from the Cataract-101 \citep{cataract101-dataset}. Using Cataract-101, \citet{edge-cataract101} explored a two-path ResNet architecture combining the RGB frame and edge detection frames to produce frame-wise, independent activity predictions. \citet{cbrcnest} developed an end-to-end model using a triplet loss consisting of activity recognition, phase recognition, and contrastive learning resulting in the state-of-the-art on the dataset.

\vspace{-10pt}
\section{Preliminaries}
In studying and bench-marking model performance, we selected combinations between 4 state-of-the-art backbone models and 4 temporal models resulting in 16 model pairs. Our model selection explore both convolution \citep{i3d,slowfast} and Transformer-based \citep{tformer,swin} backbones, as well as CNN \citep{tcn}, RNN \citep{gru}, and Transformer \citep{satt} temporal models.

\vspace{-10pt}
\subsection{Backbone Architectures}
During training of long videos, the stage-one backbone model usually learns to generate discriminative features for smaller units (either a single frame or a clip consisting of few frames) within the video. We focus on generating features for clip units due to their added ability to encapsulate the temporal motion of the objects in the scene when compared to only the spatial layout in a single frame. We explore two prominent groups of ``spatial-temporal" backbone models: 3D CNN and Transformer-based models.

\subsubsection{3D CNN-based Backbones.}
3D CNNs are an intuitive method to extend single frame learning to video clip learning. First introduced in \citep{3dc}, 3D CNNs form the basis for a family of model architectures. Two landmark architectures -- the Inflated 3D ConvNet (I3D)\citep{i3d} and the SlowFast \citep{slowfast} -- are explored in our work. 

\subparagraph{Inflated 3D ConvNets (I3D)} I3D uses a Inception-V1 2D convolution model pretrained on ImageNet cloned over the temporal dimension to train on 3D video clips.

\subparagraph{SlowFast} This model proposed by \citet{slowfast} is a two stream path for video recognition. The ``slow" path sub-samples the input clip at a low frame rate and uses large spatial convolutions to extract visual information. The ``fast" path sub-samples the input clip at a fast frame rate and uses spatially small, temporally deep convolutions to capture rapid motions. The two pathways are connected through lateral connections at each layer.

\subsubsection{Transformer-based Backbones.}
The Transformer \citep{satt} which has revolutionized the Natural Language Processing (NLP) field has two distinct capabilities when processing sequences: (1) the ability to capture long-range dependencies, and (2) incredible compute parallelization, leading to scalability. Transformers are the state-of-the-art on many tasks including machine translation and question-answering \citep{mtr,txl}. Recently, Transformers have been successfully applied to visual tasks. Vision Transformers (ViT) \citep{vit} treats images patches as input tokens to the Transformer's self-attention units. 

\subparagraph{TimeSformer} \citet{tformer} introduced the TimeSformer which extends the ViT into video understanding by performing self-attention across compact embeddings of image patches from each frame in a clip. They further investigated several self-attention designs over the space-time volume and assessed each on benchmark datasets. However, TimeSformer's computational complexity grows quadratically with image dimension and clip duration. 

\subparagraph{Video Swin Transformer} \citet{swin} proposed the Video Swin Transformer which extended Swin Transformers \citep{iswin} to introduce a TimeSformer alternative that yielded linear computational complexity growth with respect to image and time dimensions. The model performs self-attention across patches within separate image windows; then ``shifts" the window in the following layers to merge the features from the previous layer (\textit{swin}=\textit{s}hifted \textit{win}dows). In this paper, we denote the Video Swin Transformer as ``Swin".


\begin{figure*}[hbtp]
\floatconts
  {fig:act_defs}
  {\caption{The predicted activity classes for each dataset with corresponding illustrations.}}
  {\includegraphics[width=1\linewidth]{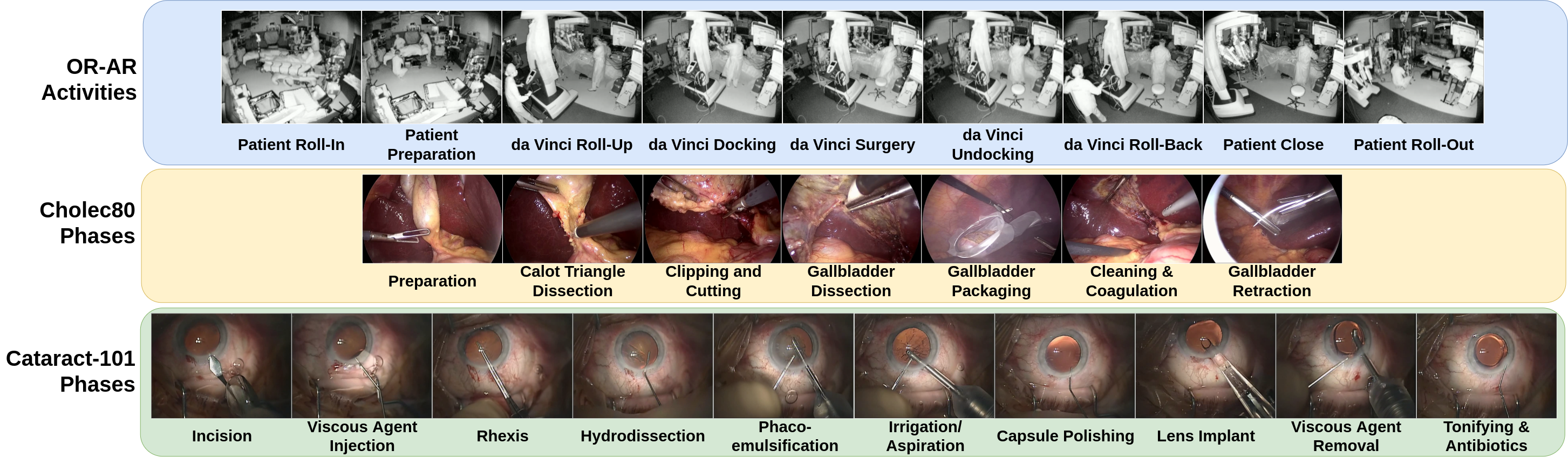}}
  \vspace{-5pt}
\end{figure*}

\subsection{Temporal Sequence Models}
A powerful backbone serves as a great foundation in many video understanding tasks; however, when dealing with videos comprising of multiple long activities, in order to obtain reliable (global) predictions, we still need to model the temporal dependencies between the extracted features. In this work, we focus on CNN, RNN, and Transformer-based temporal sequence models.

\paragraph{Temporal Convolution Networks.} \citet{tcn} introduced a new class of time-series models consisting of hierarchical temporal convolutions to capture long-range dependencies in videos and used it to perform action segmentation on videos. Assuming a low number of layers, these models are faster to train than traditional RNNs. 

\paragraph{Recurrent Neural Networks.} RNNs have been used in a wide range of applications and require little introduction. They follow a sequential nature where the output at any time depends on the ``memory" of previous inputs as well as the current input. Since the memory compactly summarizes all the previous inputs, RNNs can suffer when learning long-range dependencies. Long-Short Term Memory Networks (LSTM) \citep{lstm} alleviate this issue by introducing several gates responsible in determining what information is worth storing in the memory. Gated Recurrent Network (GRU) \citep{gru} (a simplified LSTM) have shown incredible performance in sequence modeling tasks and hence we include it in our experiments. A bi-directional GRU architecture contains a second GRU which operates on the sequence in reverse order, giving the model knowledge of future data.

\paragraph{Transformers.} Transformers are very efficient for training when the input sequence length is significantly smaller than the feature dimensions. However, this is not always the case when dealing with videos of variable length. Nevertheless, we include it \citep{satt} in our experiments. 

\section{Datasets}

\begin{figure*}[htbp]
\floatconts
  {fig:training_steps}
  {\caption{Two-stage model training process. In stage 1, the backbone model learns to classify clips into different surgical activities. In stage 2, a video is cut into N clips and passed through the backbone model to extract one feature per clip. Then, the sequence of features is used to train a temporal model to classify each clip as one of the surgical activities.}}
  {\includegraphics[scale=0.27]{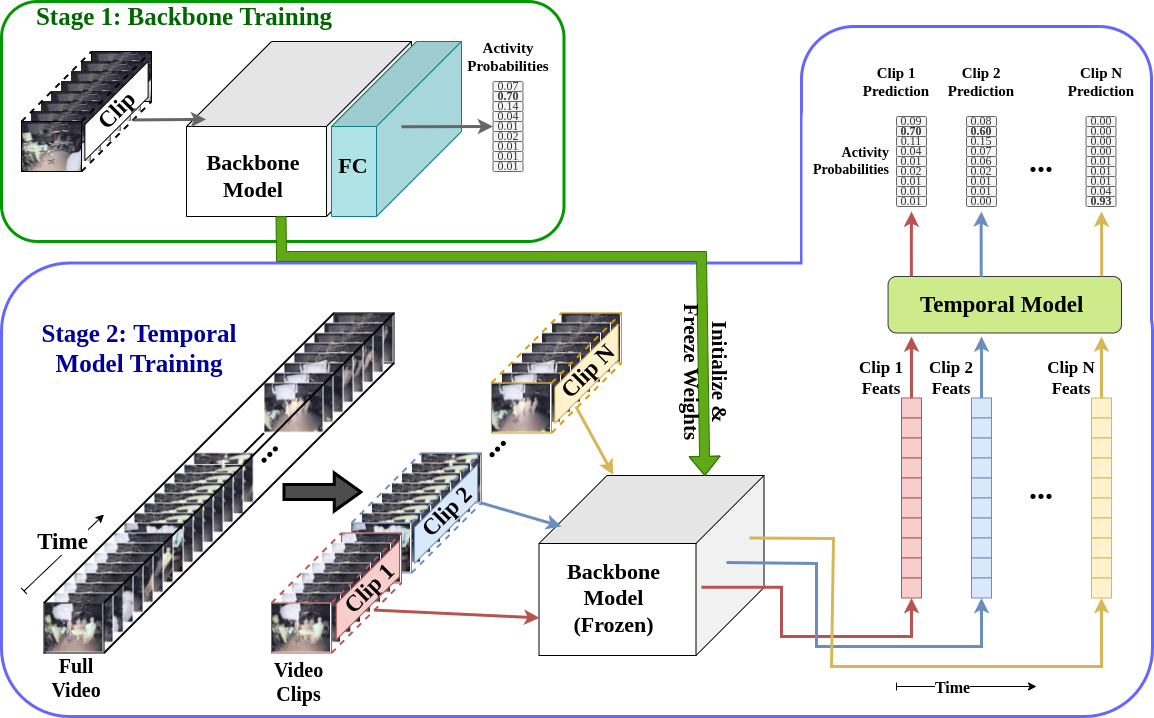}}
\end{figure*}

\paragraph{OR-AR}
We benchmarked the backbone \& temporal model combinations on three different datasets. Our first dataset, the base Operating Room Activity Recognition (OR-AR) dataset was first introduced by \citet{mic20}. It contains ~820 videos taken using time-of-flight sensors placed around the OR during robotic surgery. Each video frame is a weighted sum of intensity and depth frames. The data was collected on 27 surgeons and includes 30 types of surgeries across two ORs over a period of two years. This dataset contains an additional 66 videos taken from a second hospital which will only be used for model adaptation experiments. Video annotations assign regions of the video into 9 class shown in \figureref{fig:act_defs}.

We created 5 data splits that measure the generalizability of the models. The random split separates the dataset into an 80-20 train-test split. The OR, surgeon, procedure, and chronological splits are described in detail in the Appendix~\ref{sec:split_strats}.

\paragraph{Cholec80}
Our second dataset, Cholec80 \citep{endonet}, contains 80 endoscopic videos of cholecystectomy procedures performed by 13 surgeons and is labeled with 7 surgical phases by a senior surgeon. The surgical phases are defined in \figureref{fig:act_defs}. The dataset also includes presence annotations for 7 surgical tools which several other papers use to aid phase predictions. The videos are recorded at 25fps using a 1920x1080 resolution. Following previous works, the first 40 videos are used for training and the remaining 40 for testing. 

\paragraph{Cataract-101}
The Cataract-101 dataset contains 101 videos from cataract surgeries captured from a microscope and labelled by a senior ophthalmic surgeon in Klinikum Klagenfurt, Austria \citep{cataract101-dataset}. The start \& end times for 10 phases (defined in \figureref{fig:act_defs}) are given. Videos are recorded at 25fps using 720x540 resolution with an average 8.3 min. length. These phases are less orderly than the other datasets. Following \citet{edge-cataract101}, we randomly split the data into 73 train \& 28 test videos.

\begin{figure*}[htbp]
\floatconts
  {fig:prediction-comparison}
  {\caption{Prediction comparison of different models for one video in the OR-AR dataset. (a) The performance of various backbones with BiGRU. (b) The performance of various temporal models with the Swin backbone.}}
  {\includegraphics[scale=0.23]{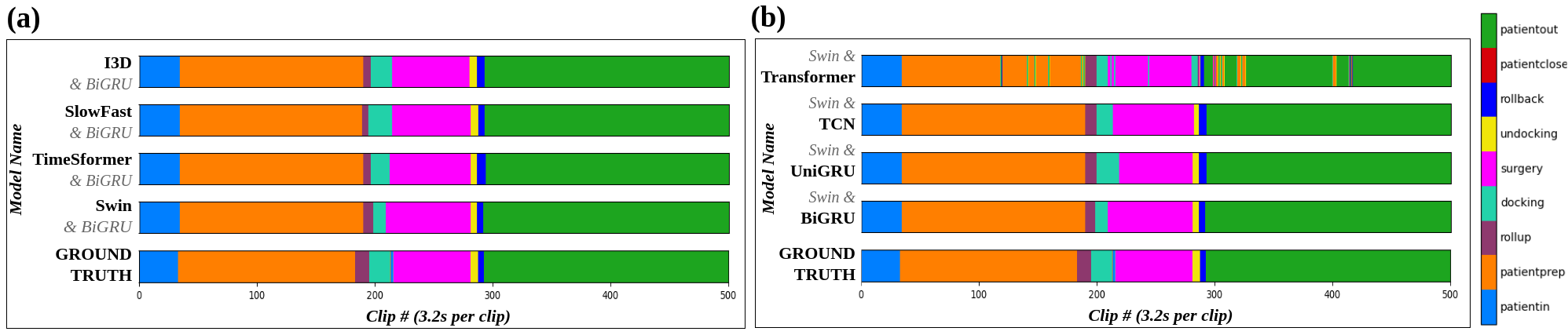}}
  \vspace{-20pt}
\end{figure*}

\subsection{Metrics.} For the Cholec80 and Cataract-101 datasets, we report the average test accuracy, precision, and recall on the epoch with the lowest training loss. On the OR-AR dataset, we report the average test-time mean average precision (mAP) score at the last training epoch.

\vspace{-5pt}
\section{Experiments} 

\paragraph{Backbone Training}
Every experiment involves two stages -- backbone training and temporal model training -- which are illustrated in detail in \figureref{fig:training_steps}.

Once the models were trained, we extracted features from the videos for each model by saving the output of the penultimate layer for each clip. For OR-AR, we noticed that dropping the clip size from 32 to 16 frames during extraction improved the performance of the subsequent temporal model. We conjecture this change increases the information density in the features resulting stronger performance.
\begin{table*}[htbp]
\caption{Average Validation mAP scores over two runs on surgical activity recognition dataset using the Random split (refer to Appendix~\sectionref{sec:split_strats} for split details).}
\label{tab:stjude-random-results-table}
\vspace{-10pt}
\centering
{%
\begin{tabular}{clcccc} 
\toprule
 \multicolumn{2}{c}{} & \multicolumn{4}{c}{\textit{Temporal Model}} \\
 \cmidrule{3-6}
 \multicolumn{2}{c}{}  & {Transformer}       & {Bi-GRU}            & {Uni-GRU}           & {TCN} \\ 
 
 \cmidrule{2-6}
 & I3D    & { 79.30$\pm$0.06 } & { 94.04$\pm$0.66 } & { 90.95$\pm$0.74 } & { 91.33$\pm$0.23 } \\ 
 \cmidrule{2-6}
 & SlowFast     & { 79.42$\pm$1.71 } & { 94.33$\pm$0.19 } & { 90.70$\pm$0.04 } & { 89.79$\pm$1.08 } \\  \cmidrule{2-6}
 \textit{Backbone}
 & TimeSformer  & { 76.23$\pm$0.33 } & { 93.20$\pm$0.04 } & { 88.89$\pm$0.66 } & { 89.59$\pm$0.07 } \\ 
 \cmidrule{2-6}
 & Swin         & { \textbf{82.50$\pm$2.35} }        & { \textbf{95.13$\pm$0.35} }       & { \textbf{92.02$\pm$0.69} }       & { \textbf{91.54$\pm$0.03} } \\
 \bottomrule
\end{tabular}}%
\vspace{-10pt}
\end{table*}

\begin{table*}[htbp]
\floatconts
  {tab:cholec-SOTA-table}%
  {\caption{Performance comparison of top bench-marked models on Cholec80. Average metrics (\%) over multiple runs along with their standard deviations.}}%
{\begin{tabular}{l|ccc} \toprule
{Model}           & {Accuracy}  & {Precision} & {Recall}    \\
\midrule
PhaseLSTM \citep{endonet} & 79.68$\pm$0.07      & 72.85$\pm$0.10      & 73.45$\pm$0.12      \\
EndoLSTM \citep{endolstm} & 80.85$\pm$0.17      & 76.81$\pm$2.62      & 72.07$\pm$0.64      \\
MTRCNet \citep{mtrcn}   & 82.76$\pm$0.01      & 76.08$\pm$0.01      & 78.02$\pm$0.13      \\
ResNetLSTM \citep{ResNetLSTM} & 86.58$\pm$1.01  & 80.53$\pm$1.59      & 79.94$\pm$1.79      \\
TeCNO \citep{tecno}       & 88.56$\pm$0.27      & 81.64$\pm$0.41      & 85.24$\pm$1.06      \\
\hline\hline
I3D+UniGRU              & 88.27$\pm$1.04      & 80.18$\pm$0.20      & 80.58$\pm$1.97      \\
SlowFast+UniGRU         & 90.47$\pm$0.46      & 83.12$\pm$2.09      & 82.33$\pm$1.22      \\
TimeSformer+UniGRU      & 90.42$\pm$0.47      & \textbf{86.05$\pm$1.13} & 83.20$\pm$1.80\\
Swin+UniGRU             & \textbf{90.88$\pm$0.01}      & 85.07$\pm$1.74      & \textbf{85.59$\pm$0.53} \\ \hline
\end{tabular}}
\vspace{-15pt}
\end{table*}

\begin{table*}[htbp]
\floatconts
  {tab:cataract101-SOTA-table}%
  {\caption{Cataract-101 performance comparison between our approach and other baselines.}}%
{\begin{tabular}{l|ccc} \toprule
{Model}           & {Accuracy}  & {Precision} & {Recall}    \\
\midrule
\citep{edge-cataract101} & 87.10      & --      & --      \\
CB-RCNeSt \citep{cbrcnest} & 96.37    & 94.89   & 94.69   \\
\hline\hline
I3D+UniGRU              & 93.69$\pm$0.21 & 91.27$\pm$0.02 & 91.05$\pm$0.41 \\
SlowFast+UniGRU         & 92.08$\pm$0.32 & 89.30$\pm$1.22 & 88.63$\pm$0.32 \\
TimeSformer+UniGRU      & 94.44$\pm$0.01 & 92.43$\pm$0.25 & 91.89$\pm$0.23   \\
Swin+UniGRU             & 94.53$\pm$0.09 & 93.05$\pm$0.09 & 91.61$\pm$0.16   \\ \hline
\end{tabular}}
\vspace{-5pt}
\end{table*}

\paragraph{Temporal Model Training.}
Once the features are extracted, each video in the training set serves as a single sample for training the temporal models. The goal of the temporal model is to learn the global logic of phase predictions across clips. Since several videos in the same batch can have vastly different lengths, we padded the shorter sequences with zeros; however, we make sure to prevent the padding from influencing the loss using a masking scheme.

\subsection{Pretraining}
To converge on the smaller Cholec80 and Cataract-101 datasets, all backbone models required pretraining on the Kinetics-400 dataset (K400) \citep{kinetics400}, a dataset consisting of ~240K training clips of human actions. For the larger OR-AR dataset, we used random initialization \citep{kaiminginit}.

\begin{table*}[htbp]
\floatconts
  {tab:model_adapt}%
  {\caption{Performance of the model adaptation techniques from hospital A to hospital B. All techniques use the Swin+BiGRU model.}}%
{\begin{tabular}{ll|cccc} \toprule
{Backbone Method}     & {Temp. Model Method}  & {mAP}  & {Accuracy} & {Precision} & {Recall}    \\
\midrule
Freeze (init: hospA) & Freeze (init: hospA)  &  72.92 & 96.34 & 53.26 & 53.66 \\%
Train (init: K400) & Train (init: random) & 83.01  & 93.47 & 68.84 & 75.99 \\
Freeze (init: hospA) & Train (init: random)   &  90.85 &  97.15  & 81.65 & 89.01 \\%
Train (init: hospA) & Train (init: random)  & 91.64  & 97.64  & 87.61 &  84.08  \\ %
\hline
\citet{or_adapt_mic22} & Train (init: random) & 88.99 & 97.26 & 86.72 & 86.02 \\ \hline %
\end{tabular}}
\end{table*}

\section{Results and Discussion}

\subparagraph{OR-AR.} The Video Swin Transformer backbone shows a clear performance edge over other backbones as illustrated in the random split results in \tableref{tab:stjude-random-results-table} and \figureref{fig:prediction-comparison}. Interestingly, the 3D CNN-based backbones (SlowFast, I3D) outperforms the attention-based TimeSformer backbone. We believe this is due to differences between the datasets. OR-AR requires a deeper understanding of motion cues because camera position is not fixed between videos and the OR scene is more complex. The TimeSformer's ``divided" spatial-temporal attention struggles against the joint spatial-temporal modelling of Swin and 3D CNN-based backbones. From our experiments, we also conclude that Swin is better at recognizing patterns across different environments. When testing on an unseen OR (refer to Appendix \tableref{tab:stjude-or-results-table}), the performance gap between Swin and other backbones is even more pronounced. Other models may rely on OR-specific visual cues resulting in lower generalizability.

\subparagraph{Cholec80.} The clip-wise models that we analyzed achieved the \textbf{new state-of-the-art on the Cholec80 dataset}. \tableref{tab:cholec-SOTA-table} demonstrates that all clip-wise models performed on-par or better than baselines from existing literature, with the Swin+UniGRU yielding the best performance in accuracy (+2.32\%) and recall (+0.35\%) while yielding strong performance in precision (+3.43\%). This is especially impressive considering several baselines utilize additional information (tool presence) to perform activity recognition. TimeSformer+UniGRU and SlowFast+UniGRU still produced strong results. Results for of all backbone \& temporal combinations are shown in Appendix  \tableref{tab:cholec-results-table}.

\vspace{-5pt}

\subparagraph{Cataract-101.} Our benchmarked models achieved strong performance on the Cataract-101 dataset (see \tableref{tab:cataract101-SOTA-table}). Our best model \textit{Swin+UniGRU} model achieved numbers slightly below SOTA (accuracy: -1.84\%, precision: -1.84\%, recall: -3.08\%). The state-of-the-art CB-RCNeSt uses additional supervision by grouping the 7 phases into 4 broader phases and training the model to predict both. Additionally, it uses a contrastive branch to differentiate visually similar frames. Since cataract surgery phases are often non-sequential, we reason that our two-stage model's global understanding may be unhelpful for this task.

\subparagraph{Backbone \& Temporal Combinations.} The results from each backbone \& temporal model combination are shown in \tableref{tab:stjude-random-results-table} for the OR-AR random split and Appendix \tableref{tab:cholec-results-table} for the Cholec80 dataset . Overall, the Swin+BiGRU combination performed the best across both datasets. Pertaining to temporal model performance, it is expected that the future-seeing models (BiGRU, Transformer) should outperform the ``online" models (UniGRU, TCN). It appears that Transformers do not meet this expectation. We believe this is due to the Transformer's complexity. The Transformer can access the global sequence from each time-step. While usually an advantage, this makes the model incredibly difficult to optimize when there is a natural order of activities in a very long sequence especially when the dataset is small and video lengths are irregular. In contrast, the BiGRU is inherently sequential and delivers strong performance.

\vspace{-5pt}
\subsection{Model Adaptation} 
We investigated the transfer learning techniques for the OR-AR task on an additional 66 recordings from another hospital (56 train, 10 test) which is 8\% the size of the OR-AR dataset. Our results are presented in \tableref{tab:model_adapt}. Using the unsupervised technique from \citet{or_adapt_mic22} to fine-tune Swin, we achieved an impressive 88.99\% test mAP (93\% of the OR-AR supervised performance). Fine-tuning using labelled data, we achieve 91.64\% mAP.  Please see Appendix~\ref{adapt} for implementation details.

\section{Conclusion}
In this paper, we focus on exploring spatial-temporal backbone architectures for surgical activity recognition and activity detection in the OR. We explore both 3D CNN-based and transformer-based architectures and show their superior performance on benchmark datasets. Furthermore, we argue that while no pretraining is required in the presence of a large labeled dataset, by benefiting from existing action recognition datasets we can overcome the performance and optimization issues that often serve as an obstacle to adoption of these models on small medical datasets. Finally, we combine the backbone models with the three predominant categories of temporal modeling and show that the Video Swin Transformer + Bi-Directional GRU produces the best performance on both Cholec80 and OR-AR datasets.

\bibliography{pmlr-sample}

\clearpage
\section*{Appendices}
\setcounter{section}{0}
\def\thesection{\Alph{section}}

\section{Implementation Details}

\subsection{Backbone Training}

To train the backbone, we selected short clips of activities from the long surgical videos; then we sub-sampled the frames within each clip according to Appendix \tableref{tab:clip-sampling} to form the input frames to the backbone. For Cholec80 and Cataract-101, we kept the scope of the backbones to 1-second long clips to match the prediction frequency of the existing works which form our baselines for comparison. We used categorical cross entropy loss on the output probabilities and used stochastic gradient descent (SGD) with momentum to update the backbone weights. We used a cosine learning rate scheduler (with Swin,Timesformer LR=0.01; SlowFast,I3D LR=0.1). We trained Swin, TimeSformer, SlowFast, and I3D for 50, 60, 196, and 100 epochs respectively for all datasets. SlowFast and I3D models also used a linear warm-up for 34 and 10 epochs respectively. The only applied augmentation technique was random flipping.

\begin{table}[htbp]
\floatconts
    {tab:clip-sampling}
    {\caption{Clip forming strategies. N(M/S) represents N-frame model inputs formed by sampling M-frame video clips by S.}}
{\begin{tabular}{lccc}
    \toprule
    \bfseries{Dataset} & \bfseries{Training} & \bfseries{Extraction} \\
    \midrule
    OR-SAR & 16(32/2) & 16(16/1) \\
    Cholec80 & 8(25/3.125) & 8(25/3.125) \\
    Cataract101 & 8(25/3.125) & 8(25/3.125) \\
    \bottomrule
\end{tabular}}
\end{table}

Backbone params, FLOPs, and performance on the OR-AR dataset can be seen in Appendix Table \ref{bb-table}. The attention-based models have higher number of parameters and FLOPs compared to their 3D ConvNet counterparts. TimeSformer has the highest number of parameters (and FLOPs) due to global connection between the image patches. Swin transformer reduces the complexity by only connecting the neighboring patches instead.

\begin{table*}[htbp]
\centering
\caption{Further information on backbone models including details about model sizes and backbone Top-1 error during testing.}
\label{bb-table}
\scriptsize
\begin{tabular}{ c | c c c c }
\hline
 {Model}       &  {\# of Params}  &  {FLOPs} & \multicolumn{2}{c}{ {Top-1 Acc. on test sets}}  \\
\cline{4-5}
                    &                       &               & Cholec80          & SAR (Random) \\
\hline
I3D                 &  27.2M                &  28.7G        & 71.36$\pm$2.19    & 85.33$\pm$0.67 \\
SlowFast            &  33.6M                &  12.7G        & 71.38$\pm$3.59    & 86.26$\pm$1.50 \\
TimeSformer         & 121.3M                & 196.1G        & 69.02$\pm$0.26    & 84.39$\pm$0.20 \\
Swin Transformer    &  88.1M                & 141.0G        & 79.09$\pm$1.29    & 86.25$\pm$2.50 \\
\hline
\end{tabular}
\end{table*}

\subsection{Temporal Model Training}

For the GRU models, we used 0.1 drop-out on input, a 64 dimension hidden state across 4 GRU layers, and a linear last layer. We report numbers for both the unidirectional and bidirectional GRU. For the TCN model, we used 7 layers with the number of channels between layers equivalent to the input feature dimension. For the Transformer model, we used an FC layer to decrease the input feature dimension to 128, followed by an 1 layer, 2 headed encoder. Lastly, we used an linear layer to obtain the final output. The Transformer was trained with 0.1 drop-out on the encoder parameters.

\subsection{Model Adaptation} \label{adapt}
For the OR-AR dataset, we explored domain adaptation techniques to understand the Swin backbone's ability to converge on an additional 66 recordings from another hospital which we randomly split into 56 training and 10 testing videos. We applied the unsupervised adaptation method proposed by \citet{or_adapt_mic22} which trains the model on strongly augmented data from the new domain using auto-generated pseudo-labels. Additionally, we fine-tuned the Swin backbone using fully-labeled data for 25 epochs starting from a model trained on the original domain. We applied a tiered learning rate with cosine decay which mostly froze lower layers (by-layer lr: [0, $5\times10^{-5}$, $1\times10^{-4}$, $2\times10^{-4}$]). For the temporal model in both techniques, we retrained the BiGRU from random under supervision for 20 epochs with a learning rate of 0.001. We found that training BiGRU from random and from a model trained on the original domain both yielded similar performance.

\subsection{Hardware.} We performed most benchmarks on four NVIDIA A100 (40GB) GPUs using the AMD EPYC 7742 64-Core CPU with 1TB RAM. Remaining experiments were conducted using four NVIDIA RTX A4000 (16GB) GPUs using the AMD Ryzen Threadripper PRO 3975WX 32-Cores CPU with 500GB RAM. 

\section{Splitting Strategies}

\subsection{OR-AR Splitting Strategies} \label{sec:split_strats}
The Operating Room Activity Recognition (OR-AR) dataset contains roughly 820 robotic surgery case-videos (.mp4) from $\approx$205 cases, on 27 surgeons and 30 types of surgeries, across two ORs, and over a period of two years. We placed four sensors around the OR to capture multiple viewpoints of each case. In order to comprehensively assess our models, we defined several train-test data splits. Each split is designed to evaluate models' generalizability across different domains.

\noindent \textbf{I.} \textbf{Random.} We randomly split the cases into a train-test split at a ratio of 80:20. Multiple views from the same case were grouped into the same set, either train or test. This splitting strategy creates homogeneous sets, and thus expresses the models' upper bound of performance.

\noindent \textbf{II.} \textbf{Operating Room.} Because the data is captured from two ORs, we can study a model's generalizability to an unseen environment by creating a train-test split on the ORs. We used data from one OR for training and use the other OR for testing. This resulted in a 63:37 train-test split.

\noindent \textbf{III.} \textbf{Surgeon.} Surgeons lead the procedure and significantly affect the surgical workflow. In this experiment, we train our model on videos from 13 surgeons and assess the model's performance on videos from the remaining 8 surgeons. The surgeons are split such that we obtain an 80\%-20\% train-test split on case videos.

\noindent \textbf{IV.} \textbf{Procedure.}  With this split, we assess if the model can generalize to novel procedures by training on 21 specific procedures (Lobectomy, Colectomy, etc.) and testing on 7 other unseen procedures (Umbilical Hernia Repair, etc.). The resulting split has a train-test split of 80:20 on case videos.

\noindent \textbf{V.} \textbf{Chronological.} We would like to see our models keep up with the evolution of OR workflow over time. We thus introduced a split where the first 80\% of chronologically-ordered cases were placed in the train set with the following 20\% in the test set.

\tableref{tab:stjude-random-results-table} in the main text shows how the models perform on Random data split. We assess the models on the remaining data splits in \tableref{tab:stjude-or-results-table,tab:stjude-surgeon-results-table,tab:stjude-surgtype-results-table,tab:stjude-timeseries-results-table} in the supplementary material (see the next page).
\begin{table*}[htbp]
\floatconts
  {tab:stjude-or-results-table}%
  {\caption{Average Validation mAP scores over 2 Trials on OR-AR Dataset using the Operating Room split}}%
{\begin{tabular}{clcccc} 
\toprule
 \multicolumn{2}{c}{} & \multicolumn{4}{c}{\it{Temporal Model}} \\
 \cmidrule{3-6}
 \multicolumn{2}{c}{}  & {Transformer}       & {Bi-GRU}            & 
 {Uni-GRU}           & {TCN} \\ 
 \cmidrule{2-6}
 & I3D          & { 49.15$\pm$2.41 }     & { 71.39$\pm$2.62 } & { 65.77$\pm$1.68 } & { 67.66$\pm$0.25 } \\ 
 \cmidrule{2-6}
 & SlowFast     & { 49.48$\pm$1.33 }     & { 72.18$\pm$0.67 } & { 68.29$\pm$0.02 } & { 67.40$\pm$0.35 } \\ 
 \cmidrule{2-6}
\textit{Backbone}
 & TimeSformer  & { 44.08$\pm$0.60 }     & { 66.80$\pm$2.28 } & { 62.44$\pm$1.05 } & { 63.88$\pm$2.57 } \\ 
 \cmidrule{2-6}
 & Swin         & { \textbf{55.41$\pm$1.40} }     & { \textbf{75.92$\pm$2.06} } & { \textbf{73.18$\pm$4.12} } & { \textbf{76.39$\pm$1.85} } \\
 \bottomrule
\end{tabular}}
\end{table*}

\begin{table*}[htbp]
\floatconts
  {tab:stjude-surgeon-results-table}%
  {\caption{Average Validation mAP scores on the last epoch over 2 Trials on OR-AR Dataset using the Surgeon split}}%
{\begin{tabular}{clcccc} 
\toprule
 \multicolumn{2}{c}{} & \multicolumn{4}{c}{\it{Temporal Model}} \\
 \cmidrule{3-6}
 \multicolumn{2}{c}{}  & {Transformer}       & {Bi-GRU}            & 
 {Uni-GRU}           & {TCN} \\ 
 \cmidrule{2-6}
& I3D          & { 67.61$\pm$1.25 } & { 88.89$\pm$1.24 } & { 84.00$\pm$1.32 } & { 83.79$\pm$0.06 } \\ 
 \cmidrule{2-6}
 & SlowFast     & { 68.66$\pm$0.83 } & { 88.81$\pm$0.91 } & { 83.78$\pm$1.56 } & { 83.68$\pm$1.18 } \\ 
 \cmidrule{2-6}
\textit{Backbone}
 & TimeSformer  & { 64.66$\pm$1.12 } & { 86.51$\pm$0.77 } & { 82.24$\pm$0.67 } & { 81.65$\pm$0.02 } \\
 \cmidrule{2-6}
 & Swin         & { \textbf{73.16$\pm$1.22} } & { \textbf{90.70$\pm$0.43} } & { \textbf{86.27$\pm$1.16} } & { \textbf{86.08$\pm$0.21} } \\
 \bottomrule
\end{tabular}}
\end{table*}
\begin{table*}[htbp]
\floatconts
  {tab:stjude-surgtype-results-table}%
  {\caption{Average Validation mAP scores on the last epoch over 2 Trials on OR-AR Dataset using the Procedure split}}%
{\begin{tabular}{clcccc} 
\toprule
 \multicolumn{2}{c}{} & \multicolumn{4}{c}{\it{Temporal Model}} \\
 \cmidrule{3-6}
 \multicolumn{2}{c}{}  & {Transformer}       & {Bi-GRU}            & 
 {Uni-GRU}           & {TCN} \\ 
 \cmidrule{2-6}
 & I3D          &  { 72.20$\pm$0.89 } &  { 91.29$\pm$0.40 } &  { 86.80$\pm$0.43 } &  { 86.61$\pm$1.46 } \\ 
 \cmidrule{2-6}
 & SlowFast     &  { 73.27$\pm$0.27 } &  { 89.96$\pm$0.63 } &  { 87.60$\pm$1.42 } &  { 86.82$\pm$0.39 } \\ 
 \cmidrule{2-6}
\textit{Backbone}
 & TimeSformer  &  { 71.21$\pm$0.06 } &  { \textbf{91.38$\pm$0.04} } &  { 85.75$\pm$0.43 } &  { 86.85$\pm$0.37 } \\
 \cmidrule{2-6}
 & Swin         &  { \textbf{77.92$\pm$2.52} } &  { 91.23$\pm$0.86 } &  { \textbf{88.98$\pm$0.37} } &  { \textbf{89.53$\pm$0.36} } \\
 \bottomrule
\end{tabular}}
\end{table*}
\begin{table*}[htbp]
\floatconts
  {tab:stjude-timeseries-results-table}%
  {\caption{Average Validation mAP scores on the last epoch over 2 Trials on OR-AR Dataset using the Chronological split}}%
{\begin{tabular}{clcccc} 
\toprule
 \multicolumn{2}{c}{} & \multicolumn{4}{c}{\it{Temporal Model}} \\
 \cmidrule{3-6}
 \multicolumn{2}{c}{}  & {Transformer}       & {Bi-GRU}            & 
 {Uni-GRU}           & {TCN} \\ 
 \cmidrule{2-6}
 & I3D          &  { 78.99$\pm$0.16 } &  { 93.65$\pm$0.63 } &  { 92.17$\pm$0.09 } &  { 89.64$\pm$0.07 } \\ 
 \cmidrule{2-6}
 & SlowFast     &  { 79.28$\pm$0.13 } &  { 92.22$\pm$1.00 } &  { 90.02$\pm$1.36 } &  { 90.97$\pm$1.32 } \\ 
 \cmidrule{2-6}
\textit{Backbone}
 & TimeSformer  &  { 76.35$\pm$1.24 } &  { 92.93$\pm$0.40 } &  { 89.72$\pm$1.02 } &  { 90.20$\pm$0.92 } \\
 \cmidrule{2-6}
 & Swin         &  { \textbf{83.35$\pm$1.70} } &  { \textbf{93.91$\pm$0.61} } &  { \textbf{92.64$\pm$0.77} } &  { \textbf{93.01$\pm$0.16} } \\
 \bottomrule
\end{tabular}}
\end{table*}
\begin{table*}[htbp]
\floatconts
  {tab:cholec-results-table}%
  {\caption{Performance comparison of different backbones and temporal models on the Cholec80 Dataset.}}%
{\begin{tabular}{ l | ccc }
\toprule
{Model}           & {Accuracy}  & {Precision} & {Recall}  \\ 
\midrule
I3D+Transformer         & 73.46$\pm$0.99      & 68.42$\pm$1.20      & 66.02$\pm$1.53 \\
I3D+TCN                 & 81.21$\pm$2.49      & 81.75$\pm$0.46      & 75.64$\pm$0.45 \\
I3D+UniGRU              & 88.27$\pm$1.04      & 80.18$\pm$0.20      & 80.58$\pm$1.97 \\
I3D+BiGRU               & 91.81$\pm$0.06      & 85.20$\pm$1.44      & 85.10$\pm$4.18 \\
\midrule
SlowFast+Transformer    & 74.12$\pm$0.03      & 69.73$\pm$0.32      & 65.99$\pm$1.15 \\
SlowFast+TCN            & 85.35$\pm$1.39      & 80.53$\pm$0.69      & 79.86$\pm$0.50 \\
SlowFast+UniGRU         & 90.47$\pm$0.46      & 83.12$\pm$2.09      & 82.33$\pm$1.22 \\
SlowFast+BiGRU          & 92.74$\pm$0.23      & 87.71$\pm$1.27      & 84.69$\pm$1.00 \\
\midrule
TimeSformer+Transformer & 73.46$\pm$0.99      & 73.93$\pm$0.32      & 68.27$\pm$2.28 \\
TimeSformer+TCN         & 87.22$\pm$0.24      & 91.75$\pm$0.46      & 81.95$\pm$0.69 \\
TimeSformer+UniGRU      & 90.42$\pm$0.47      & 86.05$\pm$1.13      & 83.20$\pm$1.80 \\
TimeSformer+BiGRU       & 92.82$\pm$1.91      & 89.70$\pm$1.34      & 86.18$\pm$2.67 \\
\midrule
Swin+Transformer        & 80.10$\pm$0.72      & 74.35$\pm$0.98      & 74.37$\pm$0.69 \\
Swin+TCN                & 87.23$\pm$0.34      & 82.57$\pm$0.75      & 82.51$\pm$0.31 \\
Swin+UniGRU             & 90.88$\pm$0.01      & 85.07$\pm$1.74      & 85.59$\pm$0.53 \\
Swin+BiGRU              & 93.87$\pm$0.04      & 89.96$\pm$0.79      & 89.65$\pm$0.58 \\
\hline
\end{tabular}}
\end{table*}
\end{document}